\documentclass[10pt,conference]{IEEEtran}
\usepackage{amsmath,amsfonts,amssymb}
\usepackage[linesnumbered,ruled]{algorithm2e}
\usepackage{amsmath}
\usepackage{array}
\usepackage[caption=false,font=normalsize,labelfont=sf,textfont=sf]{subfig}
\usepackage{textcomp}
\usepackage{stfloats}
\usepackage{url}
\usepackage{verbatim}
\usepackage{graphicx}
\usepackage{bm}
\usepackage{cite}
\usepackage{multirow}
\DeclareMathOperator*{\argmax}{argmax}
\newcommand\sbullet[1][.5]{\mathbin{\vcenter{\hbox{\scalebox{#1}{$\bullet$}}}}}

\begin{document}

\title{Solving Few-Shot Multiobjective Multitask Optimization via Iterative Sequential Transfer}

\author{
    \IEEEauthorblockN{Tingyang Wei}
    \IEEEauthorblockA{\textit{College of Computing and Data Science} \\
    \textit{Nanyang Technological University}\\
    Singapore \\
    tingyang001@e.ntu.edu.sg}
    \and
    \IEEEauthorblockN{Haofeng Wu}
    \IEEEauthorblockA{\textit{College of Computing and Data Science} \\
    \textit{Nanyang Technological University}\\
    Singapore \\
    haofeng.wu@ntu.edu.sg}
    \and
    \IEEEauthorblockN{Ananda Phan Iman}
    \IEEEauthorblockA{\textit{Department of AI Convergence} \\
    \textit{Gwangju Inst. of Sci. \& Tech. (GIST)}\\
    South Korea \\
    anandaphan@gm.gist.ac.kr}

    \and 

    \IEEEauthorblockN{Zhao Wei}
    \IEEEauthorblockA{\textit{Centre for Frontier AI Research (CFAR)} \\
    \textit{A*STAR}\\
    Singapore \\
    wei\_zhao@a-star.edu.sg}
    \and
    \IEEEauthorblockN{Jiao Liu}
    \IEEEauthorblockA{\textit{College of Computing and Data Science} \\
    \textit{Nanyang Technological University}\\
    Singapore \\
    jiao.liu@ntu.edu.sg}
    \and
    \IEEEauthorblockN{Yew-Soon Ong}
    \IEEEauthorblockA{\textit{College of Computing and Data Science} \\
    \textit{Nanyang Technological University}\\
    Singapore \\
    ASYSOng@ntu.edu.sg}
}


\markboth{Journal of \LaTeX\ Class Files,~Vol.~14, No.~8, August~2021}%
{Shell \MakeLowercase{\textit{et al.}}: A Sample Article Using IEEEtran.cls for IEEE Journals}


\maketitle

\begin{abstract}
Applying knowledge transfer across multiple optimization tasks, multitask optimization (MTO) emerges as a promising approach to solving synergistic optimization tasks simultaneously. 
However, the development of effective knowledge transfer mechanisms in MTO fundamentally relies on aligning elite solution distributions across tasks. 
This dependency creates a critical bottleneck in \emph{few-shot optimization} regimes, as restricted evaluation budgets impede the identification of elite solution distributions required for beneficial transfer. 
This challenge is exacerbated in multiobjective multitask problems, where each optimizer must approximate a continuous Pareto manifold rather than a single optimal point.
This paper introduces Iterative Sequential Transfer (IST) to circumvent this bottleneck. 
We model MTO as a sequence of sequential transfer optimization problems, concentrating evaluations on a single target per iteration. 
We propose a likelihood-informed task prioritization mechanism to maximize transfer utility by identifying the task most likely ready for knowledge integration. 
Empirical results on benchmark and real-world problems verify the effectiveness of the proposed method under tight budgets.

\end{abstract}


\section{Introduction}
\IEEEPARstart{T}{ransfer} optimization~\cite{ong-discuss} has garnered significant attention in recent years as a novel approach to optimization problems by fully leveraging inter-task relationships. 
Considering that real-world optimization problems seldom exist in isolation~\cite{feng2023evolutionary}, transfer optimization methods are designed to avoid optimizing given tasks from scratch and to alleviate the excessive computational burden. 
Several conceptual realizations of the transfer optimization paradigm, including sequential transfer optimization~(STrO)~\cite{curbing, gmm3}, multitask optimization~\cite{mfea, mfea2, pmto}, and multiform optimization~\cite{multiform}, have spawned numerous studies in the context of knowledge transfer.

In particular, multitask optimization~(MTO)~\cite{mfea, pmto} emerges as a ubiquitous approach to solving multiple optimization tasks simultaneously by exploiting the inter-task synergies.
MTO can be formulated as follows:
\begin{equation}\label{equation: s-emt}
\min f_k(\mathbf{x}_k),~~\text{s.t.} ~ \mathbf{x}_k \in \Omega_k,~~k \in \{ 1,\ldots,K \},
\end{equation}
where $\Omega_k$ is the decision space of the $k$-th optimization problem, and $f_k$ is the objective function for the $k$-th task.
MTO tackles distinct problems simultaneously, thereby aiming at generating outputs $(\mathbf{x}_1^{*}, \mathbf{x}_2^{*}, \ldots, \mathbf{x}_K^{*})$ that are the optimal solutions for each task.
To achieve this, MTO seeks to develop proper knowledge transfer mechanisms among distinct optimization problems, either implicitly sharing the solution components~\cite{mfea, mfea2, momfea2} or explicitly building mapping among task pairs~\cite{autoencoding, sysu}.
Benefiting from these knowledge transfer mechanisms, recent years have witnessed significant advances of MTO across a plethora of applications~\cite{dozen} encompassing system-in-package design~\cite{SIP}, grid shell design~\cite{wei2026}, and collision-avoidance control~\cite{avoidance}.

Albeit the surging advances, the efficacy of these transfer mechanisms fundamentally hinges on the alignment of elite solution distributions across tasks~\cite{wei}. 
For knowledge transfer to be beneficial, the source task should provide a high-quality solution distribution~\cite{dra} in its own search space to accurately guide the target task.
Otherwise, stagnant source tasks are likely to trigger detrimental negative transfer~\cite{dra}.
This dependency creates a critical bottleneck in the \emph{few-shot optimization} regime. 
Under stringent evaluation budgets, limited evaluations are typically dispersed across all tasks, failing to identify high-quality solutions within any single task and thereby giving rise to stagnated search or negative transfer~\cite{gra}.
This challenge is further intensified in the multiobjective case, where the optimizer must approximate a continuous Pareto manifold rather than a single optimal point~\cite{wei2026}. 
Dispersing a restricted budget across multiple multiobjective search tasks more likely prevents any task from identifying an elite solution set, exacerbating the risks of negative transfer and inefficient resource utilization.

To address this, we position sequential transfer optimizers as a practical approach for solving few-shot multi-objective multitask optimization, and provide empirical evidence through an Iterative Sequential Transfer (IST) framework.
Sequential transfer optimization~(STrO) can be formally denoted as:
\begin{equation}
\label{eq:seq-transfer}
\min_{\mathbf{x}_T \in \Omega_T}\ f_T(\mathbf{x}_T)\quad \text{given}\quad \{\mathcal{D}_{S_k}\}_{k=1}^{K},
\end{equation}
where $\mathcal{D}_{S_k}$ is the optimization history for the source task $S_k$ including evaluated solutions $\{(\mathbf{x}_{S_k}^{(t)}, f_{S_k}(\mathbf{x}_{S_k}^{(t)}))\}_{t=1}^{N_t}$.
In this regard, MTO can be explicitly converted into a sequence of STrO problems, where at each iteration we choose a target task $T$ and its source set $\{S_k\}$ by maximizing the universal transfer utility, so that knowledge transfer is applied only when it is most beneficial.
This can potentially alleviate the few-shot challenge.
Meanwhile, STrO has recently been developed into a fruitful line of inquiry with both theory-informed formulations~\cite{jiao_forward, Jiao_Inverse, moo-finv} and principled empirical analyses~\cite{xue_analysis}, making it a solid foundation for our IST framework.
To effectively convert the MTO into a sequence of STrO as per the search dynamics, we introduce a simple likelihood-informed task prioritization mechanism. 

In this study, we focus on the intricate multiobjective multitask optimization (MOMTO) under a stringent computational budget. 
The inherent complexities of multiobjective optimization, coupled with the demands of few-shot optimization, exacerbate the difficulty of MTO.
To verify our position, we test the IST framework based on a recent STrO method, \emph{forward-inverse transfer evolutionary multiobjective optimizer}~(F-invTrEMO)~\cite{moo-finv}, in both few-shot MOMTO benchmark and real-world problems.
Besides, we impose this IST framework on another potent STrO method, AMTEA~\cite{curbing}, showcasing the generality of the proposed method.
We position that the IST framework not only can highlight the potential for more effective MTO in few-shot multiobjective domains, but also offers a flexible foundation for blending advances in STrO and MTO.

The relevant source code can be found in the link: https://github.com/ambigeV/stro.

\begin{figure*}[!t]
\centering
\subfloat[]{\includegraphics[width=2.4in]{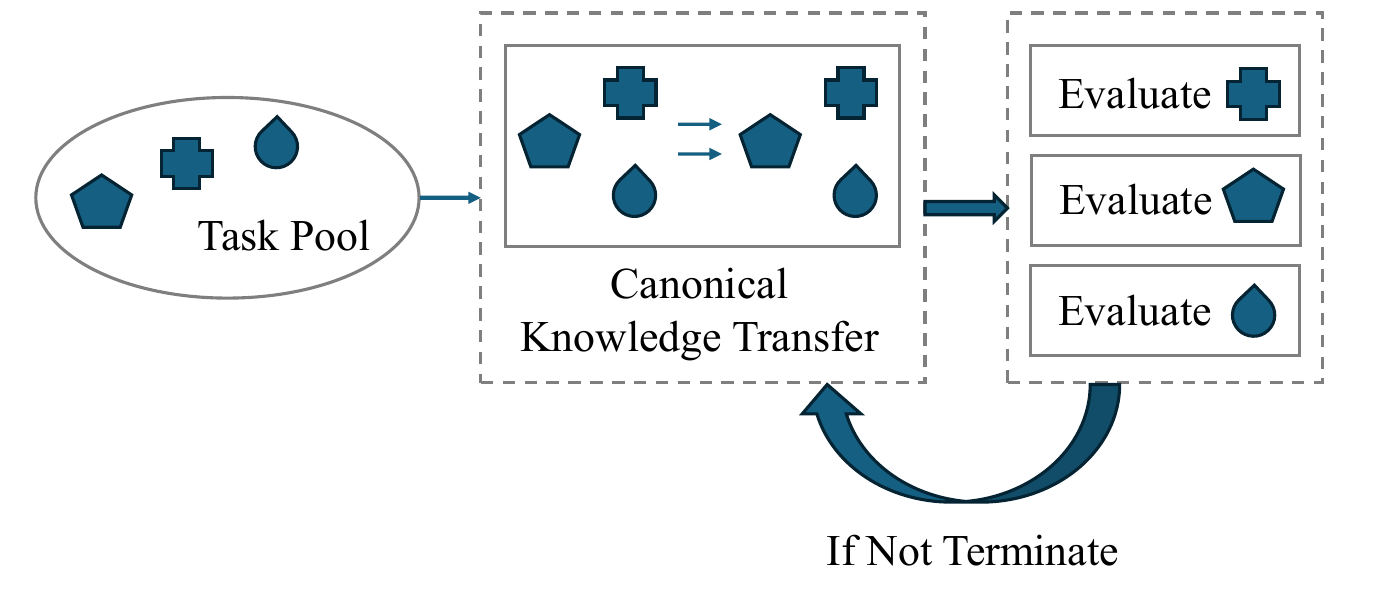}%
\label{fig_first_case}}
\hfil
\subfloat[]{\includegraphics[width=3.6in]{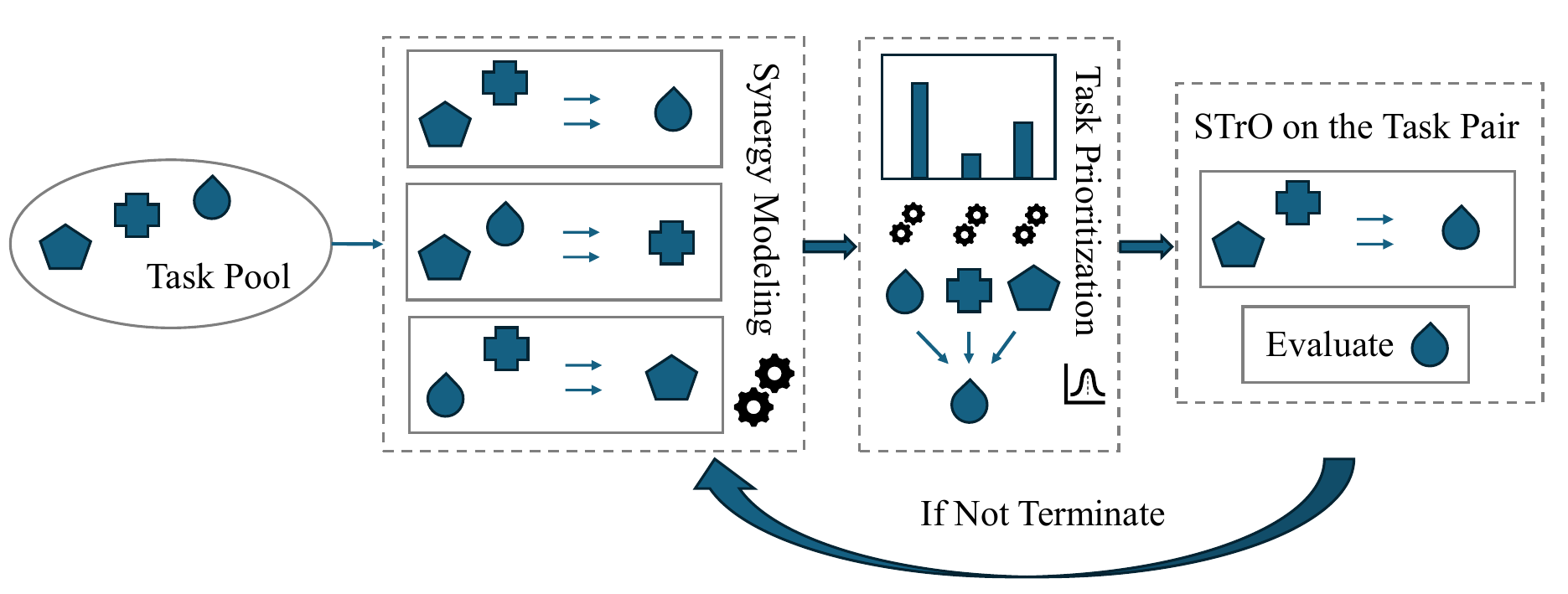}%
\label{fig_second_case}}
\caption{Workflow of distinct multitask optimization frameworks. (a) Standard multitask optimization (tasks evaluated evenly) (b) The proposed multitask optimization with iterative sequential transfer optimization (tasks evaluated selectively).}
\label{fig_sim}
\end{figure*}

\section{Related Works}\label{sec: related}
\subsection{Few-Shot Multiobjective Multitask Optimization}
This work addresses few-shot multiobjective MTO (MOMTO) by extending the scalar functions $f_k(\cdot)$ in (\ref{equation: s-emt}) to vector-valued functions $F_k(\cdot)$ with $m$ objectives. Unlike standard MTO benchmarks that allow $O(10^5)$ evaluations per task \cite{multi-benchmark}, we constrain the budget for each task to $O(10^2)$. This \emph{few-shot optimization} setting better reflects real-world constraints and demands more efficiency of search and transfer mechanisms. To our best knowledge, the research efforts in this niche area remains limited. 
However, preliminary approaches include maintaining diverse types of surrogate models~\cite{songbai} or constructing inverse multitask models~\cite{moo-finv} to tackle multiple expensive multiobjective tasks.
The latter method, F-invTrEMO~\cite{moo-finv}, serves as a baseline method in this paper, due to its flexibility for extension.

\subsection{Sequential Transfer Optimization} STrO, as formalized in (\ref{eq:seq-transfer}), leverages optimization experiences from source data or models to expedite the search in a target task. The pioneering method, AMTEA, applied transfer stacking to adaptively combine existing surrogates for target multiobjective problems \cite{curbing}. 
Other related studies focus on bridging source and target domains via transfer Gaussian Processes \cite{jiao_forward}, inverse modeling \cite{Jiao_Inverse}, or optimal transport \cite{weiming}. 
Motivated by recent theoretical progress \cite{jiao_forward, pmto, haofeng} and improvements in scalability \cite{gmm3}, we propose to solve few-shot MOMTO through iterative sequential transfer, inspired by the structural similarities between MTO and STrO and the few-shot optimization challenge mentioned in Section I.


\begin{algorithm}
\label{alg: IST}
\DontPrintSemicolon
    \caption{General Framework of IST}
    \KwData{Task size $K$, Initial budgets $N_{init}$, Total budgets for each task $N_{tot}$, Objective functions $F_k$ for each task $k$, Task prioritization function $\phi(\sbullet)$.}
    \KwResult{Optimal solutions for each optimization task.}
    \ForEach{task $k$}{
    Evaluate the objective function $F_k$ of task $k$ for $N_{init}$ iterations\;
    $Eval_k \gets N_{init}$
    }
    \While{termination condition is not met}
    {
    $k \gets \argmax_{t \in \{1, \ldots, K\} \land Eval_t < N_{tot}}\phi(t)$\;
    $Eval_k \gets Eval_k + 1$\;
    Solve formulation (\ref{eq:seq-transfer}) with task $k$ as the target task\;
    }
\end{algorithm}
\section{Multitask Optimization via Iterative Sequential Transfer}\label{sec: ist_framework} 

This section delineates the proposed IST framework. 
We first elucidate the workflow of the IST framework. 
Next, we introduce the base sequential transfer optimizer utilized in this study: F-invTrEMO. 
By integrating this optimizer within the IST framework, a likelihood-informed task prioritization mechanism is proposed. 
This approach effectively models multitask optimization as a sequence of iterative sequential transfer optimization problems.
\subsection{General Framework}
The proposed IST framework pursues the same objective as the MTO paradigm described in formulation (\ref{equation: s-emt}), that is, generating the optimal solution set $(\mathbf{x}_1^{*}, \mathbf{x}_2^{*}, \ldots, \mathbf{x}_K^{*})$. 
However, unlike standard MTO that disperses evaluations across tasks evenly, IST can model the problem as a sequence of STrO tasks as in (\ref{eq:seq-transfer}). 
Per iteration, a single target task $T$ is selected to receive an evaluation based on a likelihood-informed task prioritization mechanism, formulated as:
\begin{equation}
\begin{aligned}&\min_{\mathbf{x}_T \in \Omega_T} \quad  f_T(\mathbf{x}_T) \quad \text{given} \quad \{\mathcal{D}_{S_k}\}_{k=1}^{K} \\
& \text{s.t.} \quad  T = \argmax_{t \in \{1, \ldots, K\} \wedge Eval_t < N_{tot}} \quad \phi(t)
\end{aligned}
\label{equation: ist-prioritization}
\end{equation}
where $\phi(t)$ quantifies the utility of assigning task $t$ as the target given the current optimization history $\{\mathcal{D}_{S_k}\}$ of all tasks.

The primary distinctions of IST lie in its single-directional knowledge transfer and its selective evaluation mode. 
As illustrated in Fig.~\ref{fig_second_case}, only the prioritized target is evaluated per iteration, reflecting an inherent budget designation process that prioritizes tasks most ready for knowledge integration. 
This formulation allows IST to leverage current STrO methods for a more meticulous controlled transfer process, to potentially circumvent the few-shot challenge. 
The complete workflow of IST is detailed in \textbf{Algorithm \ref{alg: IST}}:
\begin{itemize}
    \item Initialization: Each task is evaluated for $N_{init}$ iterations, generating the source datasets $\mathcal{D}_S$. 
    \item Task Prioritization: A target task is identified per iteration by the task prioritization function $\phi(\cdot)$ to maximize transfer utility, while ensuring no task exceeds the total evaluation limit $N_{tot}$.
    \item Sequential Transfer Optimization: The identified target task is optimized by transferring knowledge from the remaining source tasks using established STrO solvers.
\end{itemize}

\subsection{Base Sequential Transfer Optimizer} 
In this study, the IST framework is instantiated using the recently proposed F-invTrEMO~\cite{moo-finv}. 
This base optimizer leverages inter-task relationships through a hybrid forward-inverse mapping approach, particularly effective for few-shot multiobjective multitask optimization.
\subsubsection{Scalarizing Multiobjective Optimization}
To handle multiple objectives within each task, the vector-valued function $F_T(\cdot)$ is scalarized using augmented Tchebycheff scalarization. For a given weight vector $\mathbf{w}$ from a $(m-1)$ dimensional simplex $\mathcal{W}$, the scalarized objective is:
\begin{equation}
\begin{aligned}
    f^{tch}_K(\mathbf{x}_K|\mathbf{w}) = &\max_{1 \leq i \leq m}\{ w_i(f_{K,i}(\mathbf{x})-(z_{K,i}^* - \epsilon)) \} + \\ &\rho \sum_{i=1}^{M} w_i f_{K,i}(\mathbf{x})
\end{aligned}
\label{equation: tch}
\end{equation}
where $z_{K, i}^*$ is the ideal point, $z_{K, i}^*-\epsilon$ provides a utopia point, and $\rho$ is a small constant to maintain Pareto optimality.

\subsubsection{Multitask Gaussian Process}
To address few-shot multi-task optimization problems, the Multitask Gaussian Process~(MTGP)~\cite{MTGP} is generally adopted to alleviate the evaluation cost and enable knowledge transfer.
Given input spaces across tasks $\Omega_k, k\in\{1,\ldots,K\}$, scalarized objective functions, $f_1^{tch}, \ldots, f_K^{tch}$, are modelled by MTGP.
During the modeling process, we have triplets $\{(i_s, \mathbf{x}_s), y_s\}_{s=1}^{N}$ with $N$ evaluated solutions, with $i_s$, the task index of the $s$-th evaluated solution, $\mathbf{x}_s \in \Omega_{i_s}$, solutions, and $y_s = f_{i_s}^{tch}(\mathbf{x}_s) + \epsilon_{i_s}$, noisy evaluations where the task-dependent noise $\epsilon_{i_s}$ is additive Gaussian noise with zero mean~(i.e., $\epsilon_{i_s} \sim \mathcal{N}(0, \sigma_{i_s}^2)$).
MTGP~\cite{MTGP} is distinct for the formulation of the multitask kernel as below:
\begin{equation}
\kappa((i, \mathbf{x}), (i', \mathbf{x}')) = \kappa_{\mathcal{T}}(i,i') \cdot \kappa_{\Omega}(\mathbf{x},\mathbf{x}')
\label{eq: MTGP}
\end{equation}
where the pair ($i, \mathbf{x}$) represents solution $\mathbf{x}$ for task $i$, $\kappa_{\mathcal{T}}$ measures the similarities among tasks, and $\kappa_{\Omega}$ measures the similarities among solutions.
Given the multitask kernel in formulation (\ref{eq: MTGP}), one can estimate the posterior distribution, $\mathcal{N}(\mu(i, \mathbf{x}), \sigma^2(i, \mathbf{x}))$, of a query pair ($i, \mathbf{x}$) as follows:
\begin{equation}
\mu_t(i,\mathbf{x}) = \bm{\kappa}_t(i,\mathbf{x})^{\intercal}(\mathbf{K}_t + \mathbf{\Lambda})^{-1}\mathbf{y}_{1:t}
\label{eq: muon}
\end{equation}
\begin{equation}
\begin{aligned}
\sigma_t^2(i,\mathbf{x}) = \kappa((i, &\mathbf{x}), (i, \mathbf{x})) - \\&\bm{\kappa}_t(i,\mathbf{x})^{\intercal}(\mathbf{K}_t + \mathbf{\Lambda})^{-1}\bm{\kappa}_t(i,\mathbf{x})
\end{aligned}
\label{eq: sigma}
\end{equation}
where $\bm{\kappa}_t(i,x) = \{\kappa((i, \mathbf{x}), (i_s, \mathbf{x}_s))\}_{s=1}^t$, $\mathbf{y}_{1:t} = \{y_s\}_{s=1}^t$, $\mathbf{K}_t = \{\kappa((i_s, \mathbf{x}_s), (i_{s'}, \mathbf{x}_{s'}))\}_{s,s'=1}^t$, and $\mathbf{\Lambda}$ is the additive noise variance matrix of MTGP.
\subsubsection{MTGP-based Forward-Inverse Transfer}
Given the scalarized objective functions in formulation (\ref{equation: tch}) and the MTGP model in formulation (\ref{eq: muon}) and (\ref{eq: sigma}) for solving few-shot optimization, the knowledge transfer can be conducted in a hybrid forward-inverse approach.
The forward mapping, $\Psi_{for}$, only approximates the scalarized objective functions, $f^{tch}_{\mathcal{T}}$, that is, $\Psi_{for}: \Omega_{\mathcal{T}} \mapsto \mathbb{R}$.
The inverse mapping, $\Psi_{inv}$, models the transformation from the $(m-1)$ dimensional simplex, $\mathcal{W}$, which includes the weight vector $\mathbf{w}$ to the solution space, $\Omega_{\mathcal{T}}$, that is, $\Psi_{inv}: \mathcal{W} \mapsto \Omega_{\mathcal{T}}$.
In this paper, we assume each task contains solutions in $d$ dimensions, that is, $\Omega_{\mathcal{T}} \subset \mathbb{R}^d$. To relieve the computational burden of multi-output GP modeling, we separated the inverse modeling into a series of single-output mapping, $\Psi_{inv, i}: \mathcal{W} \mapsto \Omega_{\mathcal{T}, i}, \Omega_{\mathcal{T}, i} \subset \mathbb{R}, i \in \{1,\ldots, d\}$. After this separate single-output GP modeling, the predictions can then be aggregated to form the original $d$-dimensional predictions.
Generally, prior to the inverse modeling process, the data $\{\mathbf{w}_s, (i_s, \mathbf{x}_s)\}_{s=1}^N$ should be prepared in advance.
In this paper, we assume that these data pairs have been constructed already, and one can refer to \cite{Jiao_Inverse} for the details and rationale behind them.
\begin{algorithm}
\label{alg: F-invTrEMO}
\DontPrintSemicolon
    \caption{Workflow of F-invTrEMO in One Pass}
    \KwData{Target task $\mathcal{T}_K$, Source tasks $\mathcal{T}_1, \ldots, \mathcal{T}_{K-1}$, Dimension size $d$, Predefined weight vector $\tilde{\mathbf{w}}$, Evaluation records $\{(i_s, \mathbf{x}_s),y_s,\mathbf{w}_s\}_{s=1}^N$, Sample size $N_S$.}
    \KwResult{Optimal solutions for the target task.}
    /*Forward MTGP Modeling*/\;
    Build a forward MTGP model, $\mathcal{N}(\mu_{fmt}(i, \mathbf{x}), \sigma^2_{fmt}(i, \mathbf{x}))$, based on optimization history, $\{(i_s, \mathbf{x}_s),y_s\}_{s=1}^N$.\;
    /*Inverse MTGP Modeling*/\;
    \ForEach{dimension $j$}{
     Build $d$ inverse MTGP models, $\mathcal{N}(\mu_{imt, j}(i, \mathbf{w}), \sigma^2_{imt, j}(i, \mathbf{w}))$, based on the evaluation records, $\{\mathbf{w}_s, (i_s, \mathbf{x}_{s,j})\}_{s=1}^N$.\;
     }
     /*Conducting Sequential Transfer Optimization*/\;
     Sample solution set $\mathcal{U}_{\mathcal{T}_K}$ from $N_S$ solutions from distribution, $\mathcal{N}(\mu_{imt}(\mathcal{T}_K, \tilde{\mathbf{w}}), \sigma^2_{imt}(\mathcal{T}_K, \tilde{\mathbf{w}}))$.\;
     Select solution $\tilde{\mathbf{x}} = \argmax\limits_{\mathbf{x}_{\mathcal{T}_K} \in \mathcal{U}_{\mathcal{T}_K}} -\mu_{fmt}(\mathcal{T}_K, \mathbf{x}_{\mathcal{T}_K})+ \beta \cdot \sigma_{fmt}(\mathcal{T}_K, \mathbf{x}_{\mathcal{T}_K})$.\;
     Evaluate solution $\tilde{\mathbf{x}}$ in the target task $\mathcal{T}_K$\;
\end{algorithm}
Considering this hybrid forward-inverse transfer mechanism, \textbf{Algorithm \ref{alg: F-invTrEMO}} demonstrates the workflow of F-invTrEMO with a single pass.
Given the datasets for both forward mapping $\Psi_{for}$ and inverse mapping $\Psi_{inv}$ per iteration, both forward and inverse MTGP models are constructed and utilized for sequential transfer optimization directly, as depicted in line 8 to line 10 in \textbf{Algorithm \ref{alg: F-invTrEMO}}.
Particularly, since this study focuses on minimization, the solution selection mechanism aims to maximize the lower confidence bound~(LCB)~\cite{GPML} as shown in line 9. 
\subsubsection{Factorized MTGP-based Forward-Inverse Transfer}
Our paper implements the MTGP-based forward-inverse transfer based on an efficient formulation in \cite{da-gp} so that the problem that, \textit{joint training MTGP can bias the source task when the volume of the source task is significantly more than that of the target task}~\cite{da-gp}, can be alleviated. 
Specifically, instead of training the MTGP model for all the $K$ tasks jointly, we replace the formulation (\ref{eq: muon}) and (\ref{eq: sigma}) with the formulation as follows:
\begin{equation}
\begin{aligned}
\mu_{fmt}(\mathcal{T}_K, &\mathbf{x}) = \sigma^2_{fmt}(\mathcal{T}_K, \mathbf{x})\{(\sum_{j=1}^{K - 1}\sigma_{\mathcal{T}_j}^{-2}(\mathcal{T}_K, \mathbf{x})\cdot\\ &\mu_{\mathcal{T}_j}(\mathcal{T}_K, \mathbf{x}) + (2-K) \cdot \sigma^{-2}_{\mathcal{T}_K}(\mathbf{x}) \cdot \mu_{\mathcal{T}_K}(\mathbf{x}) \}
\end{aligned}
\label{eq: new_muon}
\end{equation}
\begin{equation}
\sigma^2_{fmt}(\mathcal{T}_K, \mathbf{x}) = 1/\{\sum_{j=1}^{K-1}\sigma_{\mathcal{T}_j}^{-2}(\mathcal{T}_K, \mathbf{x}) + (2-K)\cdot\sigma^{-2}_{\mathcal{T}_K}(\mathbf{x})\}
\end{equation}
where $\mu_{\mathcal{T}_j}(\mathcal{T}_K, \mathbf{x})$ and $\sigma^2_{\mathcal{T}_j}(\mathcal{T}_K, \mathbf{x})$ denote the posterior mean and variance of the solution $\mathbf{x}$ in task $\mathcal{T}_K$ with MTGP using only the source task $\mathcal{T}_j$ and the target task $\mathcal{T}_K(\mathbf{x})$, $\mu^{-2}_{\mathcal{T}_K}(\mathbf{x})$ and $\sigma^{-2}_{\mathcal{T}_K}(\mathbf{x})$ are posterior mean and variance of the target task $\mathcal{T_K}$ using only the single-task GP.
This formulation is practically useful inspired by recent works in STrO domains~\cite{jiao_forward, haofeng}.

\subsection{Likelihood-Informed Task Prioritization}\label{subsec: prioritization}
The efficacy of the IST framework hinges on identifying the optimal target task $T$ to receive the next evaluation budget. 
We instantiate the task prioritization function $\phi(\cdot)$ by leveraging the inter-task relationships captured during the MTGP modeling process.

\subsubsection{Task Prioritization Function}
The search synergy between any task pair $(i, i')$ is explicitly quantified through the task kernel parameter $\kappa_{\mathcal{T}}(i, i')$, where tasks $i$ and $i'$ serve as source and target, respectively. This kernel can be estimated online per iteration as outlined in \textbf{Algorithm \ref{alg: F-invTrEMO}}. As indicated in \cite{haofeng}, for sequential transfer optimization with a single target task $\mathcal{T}_K$, a larger task search synergy parameter $\kappa_{\mathcal{T}}(\mathcal{T}_S, \mathcal{T}_K)$ suggests that Gaussian process optimization with the LCB component can achieve a lower optimization regret bound compared to source tasks with lower search synergy parameters.

Because the search synergy parameter is learned through maximum likelihood estimation, it can be treated as the statistical likelihood that a given task pair should serve as source and target, respectively~\cite{MTGP, da-gp}. 
We therefore formulate the task prioritization function as:
\begin{equation}
\phi(t) = \min_{i \neq t} \kappa_{\mathcal{T}}(i, t) .\label{eq:assignment}
\end{equation}
While high $\kappa_{\mathcal{T}}$ is generally preferred for beneficial transfer, the $\min$ operator in (\ref{eq:assignment}) ensures the selected target $t$ is compatible with the entire source task pool. 
By maximizing this lower bound, the framework prioritizes targets that maintain a high degree of synergy across all potential source tasks, thereby ensuring reliable knowledge integration.

\subsubsection{Stochastic Task Prioritization}
Given the definition of the task prioritization function $\phi(t)$, according to \textbf{Algorithm \ref{alg: IST}}, we assign the target role to task $k$ with the largest $\phi(t)$.
However, the $\argmax$ selector may cause certain tasks to remain static for too many iterations, violating the assumption in the sequential transfer optimization that all the source tasks can achieve high-quality solution distributions.
To mitigate this, we adopt the softmax function to relax this $\argmax$ selector as follows:
\begin{equation}
    \mathcal{T}_K \sim Multinomial(1; \{P_1, \dots, P_K\})
\end{equation}
\begin{equation}
    P_t = \frac{\exp(S\cdot(\max\{\phi(t)-\theta, 0\}))}{\sum_{j=1\land Eval_j<N_{tot}}^K \exp(S\cdot(\max\{\phi(j)-\theta, 0\}))}
\label{eq: softmax}
\end{equation}
where the target task $\mathcal{T}_K$ is sampled from a multinomial distribution with probabilities $P_t$ computed using this softmax function as defined in formula (\ref{eq: softmax}).
In this formulation, $S$ controls the selective pressure of the softmax function: a higher $S$ makes it more similar to $\argmax$ function, while a lower $S$ makes it resemble the uniform selection.
Moreover, the parameter $\theta$ in formula (\ref{eq: softmax}) serves as a threshold value to maintain search vitality. Specifically, while tasks with $\phi(t) > \theta$ receive an exponential boost in selection priority, all tasks falling below this threshold share an identical baseline weight ($e^0 = 1$). This ensures that even tasks with extremely low source-target likelihood remain selectable with a non-zero probability, preventing any task from being perpetually frozen during the iterative process.


\section{Experimental Studies}\label{sec: exp}
To verify the effectiveness of the proposed IST framework and the likelihood-informed task prioritization mechanism, we conduct comparative studies on MOMTO benchmark problems and multiobjective multitask hyperparameter optimization problems. 
Moreover, to further justify the generality of the framework, we instantiate another IST-based algorithm by extending a renowned sequential transfer optimizer, AMTEA~\cite{curbing}, to AMTEA-IST in Section IV.D. 
We term the F-invTrEMO implementation under the IST framework as F-invTrEMO-IST.
We compare F-invTrEMO-IST with the classical ParEGO, because many STrO methods~\cite{jiao_forward,Jiao_Inverse,moo-finv} build upon this single-task few-shot multiobjective optimizer, and the few-shot multiobjective multitask optimizer, F-invTrEMO.
One can refer to more detailed parameter settings in the supplementary materials~\footnote{Supplymentary Materials: https://zenodo.org/records/19060595}.
Experiments upon benchmark problems are independently repeated 20 trials and upon hyperparameter optimization problems are repeated 10 independent trials.


\subsection{Test Problems}
We conduct the comparative studies on multiobjective multitask benchmarks~\cite{multi-benchmark}.
It includes nine MOMTO problems and each problem contains two tasks with certain relationships regarding search space similarity and optima intersection.
The optima intersections includes complete intersection~(CI), partial intersection~(PI), and no intersection~(NI), while the search space similarity includes high similarity~(HS), medium similarity~(MS), and low similarity~(LS).
Nine sets of problems can be constructed with these attributes, including CIHS, CIMS, CILS, PIHS, PIMS, PILS, NIHS, NIMS, and NILS.
However, NIMS and NILS are not included in this paper, since our works are based on decomposition-based multiobjective optimizer, which cannot solve multitask problems with distinct objective sizes like NIMS and NILS.
As for this limitation, we leave it as a future direction.
One can refer to the supplementaries for more details on test problems.

We adopt the inverted generational distance~(IGD+)\cite{IGD} as the metric to quantify the performance of the algorithms, as recommended in~\cite{multi-benchmark}.
One can refer to the details of IGD+ in the supplementary materials.
The statistical significance test is conducted by the Wilcoxon signed-rank test.

\begin{figure*}[!t]
\centering
\subfloat[]{\includegraphics[width=2.60in]{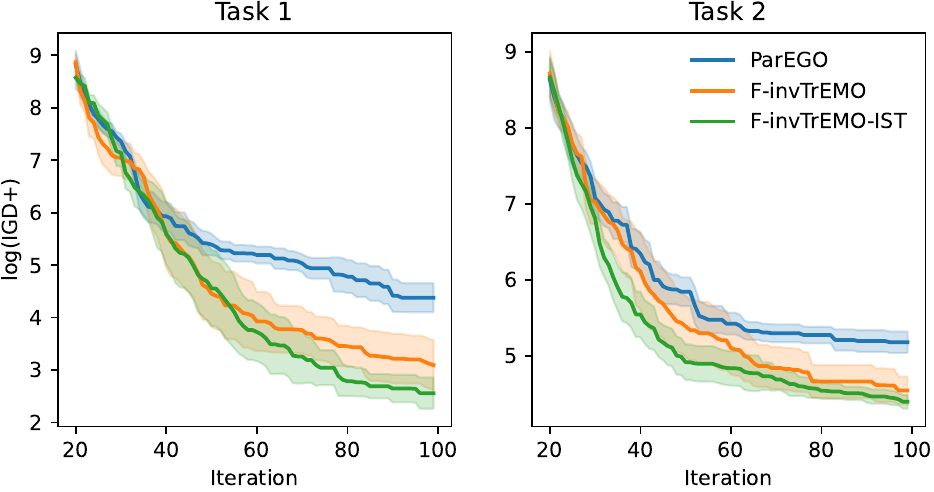}%
\label{PIHS}}
\subfloat[]{\includegraphics[width=2.60in]{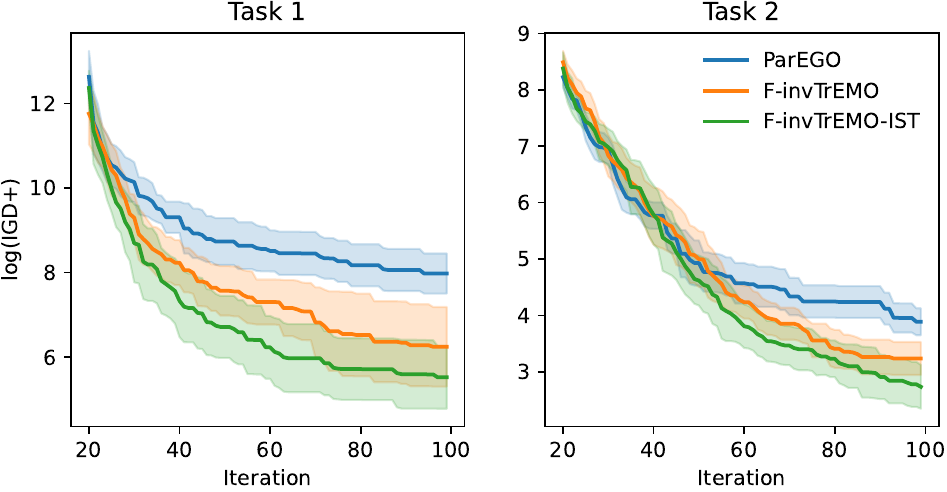}%
\label{NIHS}}
\hfil
\caption{The IGD+ convergence trends of ParEGO, F-invTrEMO, and F-invTrEMO-IST upon multitask optimization benchmark (a) PIHS (b) NIHS.}
\label{fig: Finv}
\end{figure*}

\begin{table*}[htbp]
\caption{IGD+ Comparison Results on multitask Optimization Benchmark For ParEGO, F-invTrEMO, and F-invTrEMO-IST}
\centering
\resizebox{0.6\textwidth}{!}{
\begin{tabular}{c|c|c|c|c}
\hline
Problems & Tasks & ParEGO & F-invTrEMO & F-invTrEMO-IST \\ \hline\hline
\multirow{2}*{CIHS} & Task-1 & 1.516E+02 (1.545E+01) + & 8.061E+01 (1.534E+01) + & \textbf{7.156E+01 (1.142E+01)} \\ \cline{2-5}
 & Task-2 & 4.877E+00 (3.150E-01) + & 3.557E+00 (2.329E-01) $\approx$ & \textbf{3.389E+00 (1.925E-01)} \\ \hline\hline
\multirow{2}*{CIMS} & Task-1 & 2.523E+02 (2.057E+01) + & 2.086E+02 (2.658E+01) + & \textbf{1.954E+02 (4.069E+01)} \\ \cline{2-5}
 & Task-2 & 4.174E+00 (2.983E-01) + & 3.819E+00 (2.859E-01) + & \textbf{3.606E+00 (1.799E-01)} \\ \hline\hline
\multirow{2}*{CILS} & Task-1 & 4.564E+01 (2.021E+00) + & 4.435E+01 (1.592E+00) + & \textbf{4.113E+01 (1.601E+00)} \\ \cline{2-5}
 & Task-2 & \textbf{3.050E-01 (1.040E-02) -} & 7.938E-01 (5.305E-02) + & 6.856E-01 (4.870E-02) \\ \hline\hline
\multirow{2}*{PIHS} & Task-1 & 8.728E+01 (9.112E+00) + & 3.234E+01 (8.086E+00) + & \textbf{1.399E+01 (1.978E+00)} \\ \cline{2-5}
 & Task-2 & 1.841E+02 (1.678E+01) + & 9.913E+01 (1.033E+01) + & \textbf{8.109E+01 (3.761E+00)} \\ \hline\hline
\multirow{2}*{PIMS} & Task-1 & 2.678E+00 (1.539E-01) + & \textbf{4.324E-01 (3.790E-02) -} & 4.784E-01 (5.205E-02) \\ \cline{2-5}
 & Task-2 & 3.736E+02 (1.269E+01) + & 3.503E+02 (5.258E+00) $\approx$ & \textbf{3.407E+02 (4.165E+00)} \\ \hline\hline
\multirow{2}*{PILS} & Task-1 & \textbf{1.008E+00 (1.390E-02) $\approx$} & 1.029E+00 (1.165E-02) $\approx$ & 1.041E+00 (9.500E-03) \\ \cline{2-5}
 & Task-2 & 1.423E+01 (6.948E-01) + & 1.258E+01 (8.887E-01) + & \textbf{1.053E+01 (6.600E-01)} \\ \hline\hline
\multirow{2}*{NIHS} & Task-1 & 3.174E+03 (6.287E+02) + & 1.651E+03 (5.881E+02) + & \textbf{5.934E+02 (2.174E+02)} \\ \cline{2-5}
 & Task-2 & 5.820E+01 (4.852E+00) + & 2.834E+01 (3.824E+00) + & \textbf{1.903E+01 (4.056E+00)} \\ \hline
\end{tabular}}
\end{table*}

\subsection{Results}
In our comparative study upon the multitask optimization benchmark, we compare the proposed method F-invTrEMO-IST with the single-task method ParEGO and the plain method F-invTrEMO without IST settings.
It can be indicated in TABLE I that both F-invTrEMO and F-invTrEMO-IST can significantly outperform the baseline method ParEGO due to the ability to exploit intertask relationship information through the forward-inverse transfer mechanism.
The exceptions occur in both problem sets with the relationship LS (i.e., CILS and PILS), where the search space between tasks has the least similarity.
Since both F-invTrEMO and F-invTrEMO-IST maintain the solution distribution by mapping the predefined weight vector $\mathbf{w}$ to the solution space, and the dataset used to train the mapping process is constrained according to the previous optimization process, the solution distribution can be spuriously biased towards the local Pareto front, hindering the subsequent optimization process.
Moreover, the least similar search space between tasks makes it difficult to utilize the knowledge transfer to help escape the local optima in the target task from source tasks.
In terms of the comparison between F-invTrEMO and F-invTrEMO-IST, it can be found in TABLE I that F-invTrEMO-IST that is implemented with IST settings can outperform the counterpart in 12 of 14 tasks, suggesting the superiority of the proposed IST framework.
In most problem sets, IST can identify the more proper source-target pair to conduct the knowledge transfer process using the proposed task prioritization mechanism, mitigating the potential negative transfer.
Since the knowledge transfer process for source-task pairs with lower transfer utilities should be suppressed in certain stages, IST can also be viewed as an approach to implement adaptive resource allocation on the fly.

\subsection{A Case Study on Real-world Application: Multiobjective Multitask Hyperparameter Optimization}
Hyperparameter optimization (HPO) is a standard topic in the field of machine learning. 
The configuration of hyperparameters for machine learning models can impact the model performance, computational resources, and interpretability to decision-makers.
In this paper, we apply F-invTrEMO-IST to optimize the hyperparameters of these models for the aforementioned criterion across distinct tasks. We consider the following two scenarios:
\begin{itemize}
    \item (\textit{HPO-1})~The first scenario contains three HPO problems. The three problems tune the hyperparameters of the same model, Random Forest, but on three distinct classification tasks: credit approval, medical diagnosis, and speech recognition problems.
    \item (\textit{HPO-2})~The second scenario contains two HPO problems. The two optimization problems tune the hyperparameters of distinct but related models on the same classification task, speech recognition problem~\cite{yahpo}.
One can refer to the definition of each objective function in supplementary materials.
\end{itemize}
\begin{table}[htbp]
\label{tab: ATP-HPO}
\caption{IGD+ Comparison Results on multitask Hyperparameter Tuning Problems for F-invTrEMO and F-invTrEMO-IST}
\centering
\begin{tabular}{c|c|c|c}
\hline
\multicolumn{1}{l|}{Problems} & \multicolumn{1}{l|}{Tasks} & F-invTrEMO      & F-invTrEMO-IST         \\ \hline\hline
\multirow{2}{*}{HPO-1}         & Task-1                     & 0.0502 (0.0028) $+$ & \textbf{0.0455 (0.0021)} \\ \cline{2-4} 
                               & Task-2                     & 0.0439 (0.0020) $\approx$ & \textbf{0.0423 (0.0014)} \\ \hline\hline
\multirow{3}{*}{HPO-2}         & Task-1                     & 0.0325 (0.0017) $+$ & \textbf{0.0303 (0.0031)} \\ \cline{2-4} 
                               & Task-2                     & 0.0282 (0.0019) $\approx$ & \textbf{0.0279 (0.0014)} \\ \cline{2-4} 
                               & Task-3                     & 0.0421 (0.0030) $+$ & \textbf{0.0392 (0.0023)} \\ \hline\hline
\end{tabular}
\end{table}

As illustrated in TABLE II, with the IST framework, the proposed F-invTrEMO-IST can outperform F-invTrEMO across all the optimization tasks, generating better machine learning model sets trading off across model precision, model size, and interpretability.
Importantly, in real-world cases, it is rare that the solution optima and search spaces share as high commonalities as benchmark problems such as CIHS or CIMS in TABLE I.
This heterogeneity in both optimal distribution and search space similarity makes it important to consider the utilization of computing resources and knowledge transfer direction, where the proposed \textit{iterative sequential transfer} framework matters.

\subsection{A Case Study on Generality of the Framework: IST based on AMTEA}
To verify the generality of the proposed IST on other sequential transfer optimizers, IST is integrated with another evolutionary sequential transfer optimizer, AMTEA~\cite{curbing}.
We term AMTEA with IST settings as AMTEA-IST.
AMTEA is originally designed for the canonical algorithms requiring thousands of function evaluations, while we integrate AMTEA into the \emph{few-shot multiobjective} multitask optimization.
To this end, we transform the multiobjective problem into single-objective problems via (\ref{equation: tch}) at each iteration. 
The single-objective optimization is then evaluated using a Gaussian Process surrogate model, with solutions sampled from the solution distribution managed by AMTEA, similar to the approach used in ParEGO and \textbf{Algorithm \ref{alg: F-invTrEMO}}.
One can refer to the supplementary materials for more details.

The results can be found in TABLE S-I in supplementary materials that both AMTEA and AMTEA-IST can outperform the ParEGO counterpart in 13 of 14 tasks, showcasing the effectiveness of the knowledge transfer in the context of few-shot MOMTO.
Comparing AMTEA to AMTEA-IST, with the meticulous transfer direction configuration, AMTEA-IST can outperform AMTEA in 11 of 14 multiobjective tasks.
Generally.
Moreover, the IST framework can facilitate a better search process for the base optimizer in multi-task settings with lower search space similarity, as shown in TABLE S-I.
The improvements upon AMTEA for CIHS and CIMS are not significant since high-quality solutions from both tasks are similar, so that the task prioritization components cannot precisely capture the inter-task relationship, thereby misleading the IST framework.
In contrast, for those problem sets with lower similarity in TABLE S-I, AMTEA-IST generally can significantly outperform AMTEA in light of the ability to filter out the least likely task pair.


\section{Conclusion}\label{sec: conclude}
In this work, we solve the few-shot MOMTO by introducing the Iterative Sequential Transfer (IST). 
Unlike conventional methods, our IST framework can model the multitask problem as a sequence of sequential transfer optimization problems. 
To facilitate this, a likelihood-informed task prioritization mechanism is devised to actively control the knowledge transfer direction.
This novel framework alleviates the inherent limitations of simultaneous evaluation processes in traditional MTO, thereby mitigating negative transfer and enhancing efficiency in few-shot regimes. 
The IST framework not only advances traditional MTO but also accommodates existing sequential transfer optimization algorithms. 
By bridging the gap between multitask and sequential transfer studies, this approach paves the way for more efficient algorithmic designs. The comprehensive evaluation across benchmark and real-world hyperparameter testbeds demonstrates the effectiveness of the proposed framework under stringent evaluation budgets.

\section*{Acknowledgments}
This research is partly supported by the National Research Foundation, Singapore and DSO National Laboratories under the AI Singapore Programme (AISG Award No.: AISG2-
GC-2023-010, Design Beyond What You Know: Material-Informed Differential Generative AI (MIDGAI) for Light-Weight High-Entropy Alloys and Multi-functional Composites (Stage 1b), the A*STAR Catalyst Project for Artificial Intelligence in Drug Discovery (AIDD) Programme (Grant No. H25A1N0004), the Centre for Frontier AI Research (CFAR) under Agency for Science, Technology and Research (A*STAR), and the College of Computing and Data Science, Nanyang Technological University.

\bibliographystyle{IEEEtran}
\bibliography{bibfiles_async}
\end{document}